\documentclass[10pt]{article}

\usepackage{fullpage}
\usepackage{setspace}
\usepackage{parskip}
\usepackage{titlesec}
\usepackage[section]{placeins}
\usepackage{xcolor}
\usepackage{breakcites}
\usepackage{lineno}
\usepackage{hyphenat}
\usepackage{makecell}

\PassOptionsToPackage{hyphens}{url}
\usepackage[colorlinks = true,
            linkcolor = blue,
            urlcolor  = blue,
            citecolor = blue,
            anchorcolor = blue]{hyperref}
\usepackage{etoolbox}


\renewenvironment{abstract}
  {{\bfseries\noindent{\abstractname}\par\nobreak}\footnotesize}
  {\bigskip}

\titlespacing{\section}{0pt}{*3}{*1}
\titlespacing{\subsection}{0pt}{*2}{*0.5}
\titlespacing{\subsubsection}{0pt}{*1.5}{0pt}

\usepackage{authblk}

\usepackage{graphicx}
\usepackage[space]{grffile}
\usepackage{latexsym}
\usepackage{textcomp}
\usepackage{longtable}
\usepackage{tabulary}
\usepackage{booktabs,array,multirow}
\usepackage{amsfonts,amsmath,amssymb}
\providecommand\citet{\cite}
\providecommand\citep{\cite}

\newif\iflatexml\latexmlfalse

\AtBeginDocument{\DeclareGraphicsExtensions{.pdf,.PDF,.eps,.EPS,.png,.PNG,.tif,.TIF,.jpg,.JPG,.jpeg,.JPEG}}

\usepackage[utf8]{inputenc}
\usepackage[english]{babel}

\usepackage{color}

\begin{document}

\title{Advancing from Automated to Autonomous Beamline by Leveraging  Computer Vision}

\author[1]{Baolu Li$^{\dagger}$}
\author[1]{Hongkai Yu$^{\dagger}$}
\author[1]{Huiming Sun}
\author[1]{Jin Ma}
\author[2]{Yuewei Lin}
\author[3]{Lu Ma$^{*}$}
\author[3]{Yonghua Du%$^{*}$
\thanks{$\dagger$ indicates equal contributions. * Co-corresponding authors: Lu Ma (email: luma@bnl.gov), Yonghua Du (e-mail: ydu@bnl.gov).} }

\affil[1]{Department of Electrical and Computer Engineering, Cleveland State University, Cleveland, OH 44115, USA.}%
\affil[2]{Computing and Data Sciences, Brookhaven National Laboratory, Upton, NY 11973, USA.}
\affil[3]{National Synchrotron Light Source II, Brookhaven National Laboratory, Upton, NY 11973, USA.
}%

\vspace{-1em}

\begingroup
\let\center\flushleft
\let\endcenter\endflushleft
\maketitle
\endgroup

\selectlanguage{english}
\begin{abstract}
{The synchrotron light source, a cutting-edge large-scale user facility, requires autonomous synchrotron beamline operations, a crucial technique that should enable experiments to be conducted automatically, reliably, and safely with minimum human intervention. However, current state-of-the-art synchrotron beamlines still heavily rely on human safety oversight. To bridge the gap between automated and autonomous operation, a computer vision-based system is proposed, integrating deep learning and multiview cameras for real-time collision detection. The system utilizes equipment segmentation, tracking, and geometric analysis to assess potential collisions with transfer learning that enhances robustness. In addition, an interactive annotation module has been developed to improve the adaptability to new object classes. Experiments on a real beamline dataset demonstrate high accuracy, real-time performance, and strong potential for autonomous synchrotron beamline operations.}\\%
\end{abstract}%

\section*{I. Introduction}\label{intro}

Synchrotron radiation, a cutting-edge technique based on synchrotron light source for studying materials from the atomic scale to macroscale, has been driving advancements in many fields, including quantum materials, microelectronics, energy storage, catalysis, biology, medicine, and environmental science in the past few decades~\cite{willmott2019introduction}. A synchrotron light source is a large-scale instrument, typically consisting of an electron storage ring with a circumference of a few hundred to thousand meters, with Synchrotron radiation beamlines installed along the ring. These beamlines extending from tens to hundreds of meters, serve as the terminals to conduct experiments for users. Given the high operation cost of synchrotron light source, efficient beamline operation has been an important topic for a long time. Moreover, in recent years, the application of artificial intelligence (AI) and machine learning (ML) in synchrotron data interpretation has significantly increased the demand for large, high-quality datasets, further increasing demand~\cite{doerk2023autonomous,pithan2023closing,maffettone2022delivering,slautin2025materials}. 

Automated sample exchange and data collection can greatly improve the beamline operation efficiency, which has been widely implemented at x-ray diffraction beamlines, x-ray scattering beamlines, and macromolecule beamlines. In addition to automation, automatic beamline~\cite{mangold2018fully,schneider2022amx,nash2022combining} optimization~\cite{xi2015general,xi2017ai,liu2024intelligent} and real-time data analysis help maintain reliable beamline operation and data quality~\cite{bicer2017real,basu2019automated}. However, a critical challenge remains: preventing collisions between beamline equipment, such as sample holders, sample stages, detectors, and robotic arms, during automated measurements. Unforeseen issues, such as fallen sample holders, stage mechanical/control failure, and oversized samples/in-situ cells, can lead to such collisions. Existing safety solutions, such as limit switches, predefined motion constraints, sensors, are static and lack adaptability to dynamic, real-time conditions.

An autonomous beamline must satisfy three essential criteria: automation, reliable data collection, and safe automated operation. Among these, ensuring safe automated operation remains the primary bottleneck to achieving autonomy. To address this challenge, we propose a computer vision-based system that integrates deep learning with multi-view cameras for real-time collision detection.

Computer vision (CV) enables machines to analyze and interpret visual information from the real world, allowing object recognition, tracking, and analysis~\cite{terven2023comprehensive,ren2015faster} through image processing and computational techniques. CV has been widely applied in object detection, recognition, and tracking~\cite{terven2023comprehensive,gong2018overview}, as well as in industrial automation and safety monitoring~\cite{kazemian2019computer}. Murphy et al.~\cite{murphy2010head} demonstrated the role of CV in predicting hazards in dynamic environments, Tian et al.~\cite{tian2020computer} reviewed its application in agricultural automation, while Dong et al.~\cite{dong2021review} applied CV techniques to structural health monitoring. At synchrotron beamlines, CV systems can identify potential hazards in real time, ensuring operational safety in complex experimental setups. By replacing human vision with cameras and algorithms, computer vision enables applications such as object detection, segmentation, and distance measurement. When combined with deep learning, it enhances real-time decision-making, enabling the system to function independently and reliably in dynamic environments.

\begin{figure}[!t]
\centering
\includegraphics[width=0.5\columnwidth]{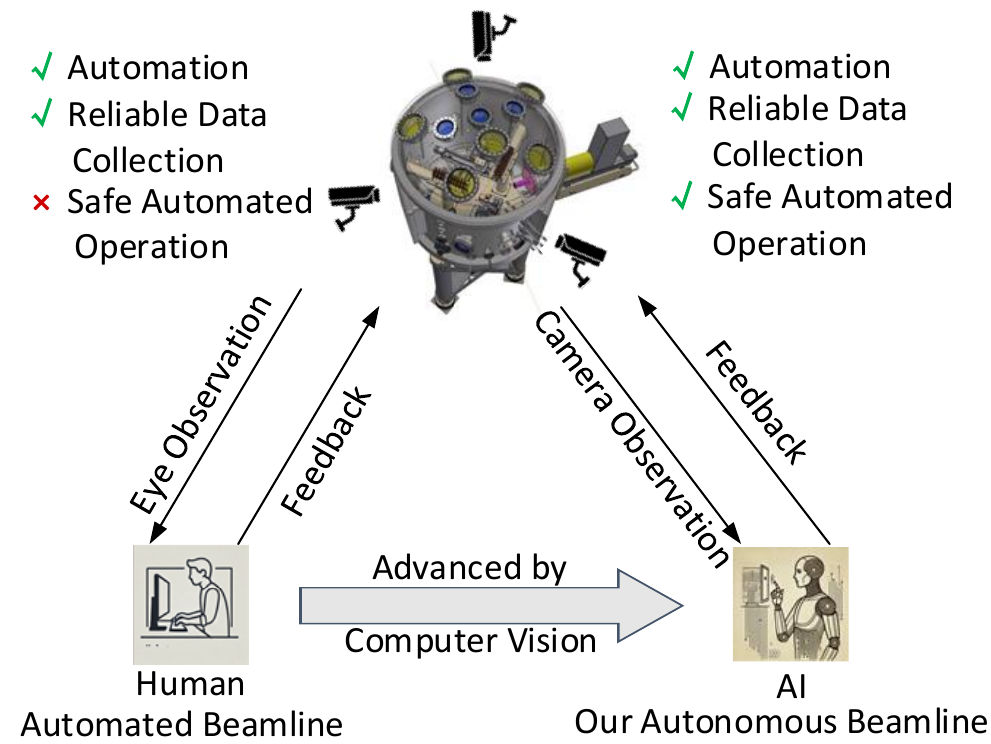}
\caption{\textbf{Comparison of automated and autonomous beamline systems.} The automated beamline relies on human eye observation for collision avoidance, whereas the autonomous beamline integrates multi-view camera to enable real-time monitoring and intelligent decision-making, significantly enhancing operation safety.}
\vspace{-1em}
\label{fig:motivation}
\end{figure}

In this effort, we developed a beamline simulation platform at Cleveland State University lab using a flexible robot arm and conducted transfer learning to adapt this technique from the lab to the real  synchrotron radiation beamline at Brookhaven National Laboratory. The final tests were carried out at 7-BM of NSLS-II. The test results show that the proposed computer vision-based system using multi-view cameras could accurately detect the collision in real time, leading to a safe autonomous beamline system.

\par\null

\section*{II. Method}\label{Method}

\begin{figure}[!t]
\centering 
\includegraphics[width=0.7\columnwidth]{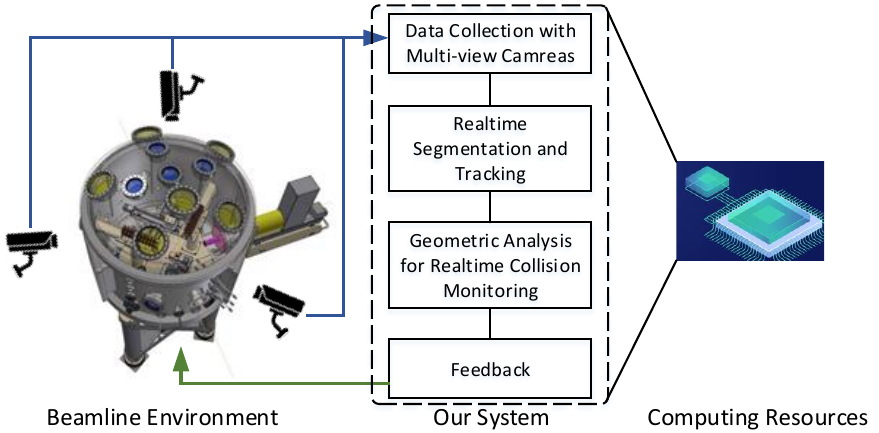}%
\caption{\textbf{High-Precision Beamline Configuration in Brookhaven National Laboratory.}}
\label{fig:beamline-config}
\end{figure}

\subsection*{2.1 Overview of Proposed System}  
As shown in Fig.~\ref{fig:beamline-config}, the proposed system mainly consists of four modules of 1) Data Collection with Multi-view Cameras; 2) Realtime Segmentation and Tracking; 3) Geometric Analysis for Realtime collision monitoring; 4) Feedback. Multi-view cameras refer to the use of multiple cameras positioned at different angles to monitor an experiment. The advantage lies in overcoming the limitations of single-view monitoring, such as blind spots, providing a broader and more comprehensive perception range. A deep learning-based method is implemented to recognize and track multiple beamline equipment in real time, incorporating geometric priors as regularization to improve accuracy and robustness. Deep learning, a subset of machine learning, leverages multi-layered neural networks to automatically extract hierarchical features. Based on the segmentation and tracking results, the system calculates the distances between pairs of equipment, enabling real-time detection of potential collisions. If the nearest distance between any two pieces of equipment falls below a predefined threshold, the system issues a warning to the beamline control system. The beamline control system can carry out proper actions to avoid this collision. We further describe the details of our proposed system in the next sections.

\subsection*{2.2 Computer Vision Methodology}\label{sec:method}
The core of the transition from automated to autonomous beamline operations is underpinned by an advanced computer vision framework that integrates segmentation, tracking, and collision monitoring. Given the video streams captured from multiple cameras viewing the same scene from different perspectives, the system uses YOLOv8~\cite{yolov8_ultralytics} for multi-equipment segmentation and AOT (Associating Objects with Transformers)~\cite{yang2021associating} for multi-equipment tracking.  Segmentation and tracking provide a stable understanding of the position and location of each piece of equipment in the beamline over time. 
Once segmentation and tracking are completed, the system computes the minimum distance between the contours of different equipment based on their geometric boundaries. If the distances between objects across all camera views are below a threshold, a potential collision is detected, leading to a triggered warning. We further describe the segmentation, tracking, geometric analysis, transfer learning, and interactive annotation in details.

\subsection*{2.2.1 Multi-Equipment Segmentation with Geometry Regularization}
To obtain the accurate equipment location, we utilize computer vision based segmentation to recognize each equipment in the camera view. Since the beamline operational environment is pre-defined, we propose a geometry regularization based on relative spatial location. We treat each interested equipment in the beamline environment as an object in computer vision. The system first applies the object instance segmentation model YOLOv8~\cite{yolov8_ultralytics} to simultaneously segment all object instances across multiple classes in each input video frame. 
Segmentation involves dividing an image into multiple segments to locate and classify objects within the scene. Specifically, an instance segmentation model identifies and delineates each object, producing precise boundaries for each piece of equipment.
Assume that the video streams from $N$ cameras are defined as $\mathbf{V}_i = \{ \mathbf{F}_{i,t} \mid t = 1, 2, \ldots, T \},  i = 1, 2, \ldots, N $, where $\mathbf{V}_i$ represents the video stream captured by the $i$-th camera and $\mathbf{F}_{i,t}$ is the frame captured by camera $i$ at time $t$. The YOLOv8~\cite{yolov8_ultralytics} model $\mathcal{S}$ processes each frame $\mathbf{F}_{i,t}$ to generate a instance segmentation mask $\mathbf{M}^{(c)}_{i,t}$ for the object class $c$, defined as:

\begin{equation}
\mathbf{M}^{(c)}_{i,t} = \mathcal{S}(\mathbf{F}_{i,t}),     
\label{eq:seg}
\end{equation}
where $c = 1, 2, \ldots, C$ and $C$ is the total number of object classes in the scene. Each mask $\mathbf{M}^{(c)}_{i,t}$ is a binary image where the foreground pixels with a value of 1 represent the region of the object class $c$.

Because the beamline operational environment is a pre-defined closed experimental environment, the relative spatial location of different object classes is a pre-known prior knowledge. For example, the class Holder should be always in the top of the class Stage in the front view. Let us define $\mathbf{G}(,)$ to be the function to obtain the relative spatial location of two object classes. Thus, the segmentation masks of Eq.~\ref{eq:seg} are  further constrained with the following regularization:  

\begin{equation}
\mathbf{G}(\mathbf{M}^{(c_1)}_{i,t}, \mathbf{M}^{(c_2)}_{i,t}) == \mathbf{G_i}(c_1, c_2),     
\label{eq:seg-reg}
\end{equation} 
where $\mathbf{G_i}(c_1, c_2)$ is the relative spatial location of two object classes $c_1, c_2$ in the $i$-th camera of the prior knowledge. The Eq.~\ref{eq:seg-reg} is hereafter referred to as \textit{geometry regularization} in this paper. We monitor the segmentation in the image sequence. If the segmentation is largely changed in adjacent frames, we will use the segmentation in the last frame with reasonable geometry regularization (Eq.~\ref{eq:seg-reg}) for the current frame. With this spatial temporal consistency, we reduce the risk of missing segmentation of the instance segmentation model. After obtaining the equipment segmentation in each frame, we then smooth the segmentation over time with the help of tracking.

\subsection*{2.2.2 Multi-Equipment Tracking}
Object Tracking refers to the process of maintaining the identity of objects across a sequence of video frames, ensuring consistent identity even as objects move or change perspective. With the help of tracking, the system will stabilize and smooth the segment of each equipment over time, therefore making the collision warning more accurate. Our system uses spatial-temporal consistency to make the object instance segmentation mask boundaries smoother and more accurate, which is realized by the multi-object tracking in computer vision.  The multi-equipment tracking algorithm will also keep the object instance mask identity consistent across multiple frames in different views, making the association easy in some potential collision scenarios. 

Our multi-view camera monitor needs to know the matching object index in the cameras in different views. We model it as a tracking-by-detection problem in computer vision. After each equipment/object is segmented/detected in each camera, the system applies the AOT (Associating Objects with Transformers)~\cite{yang2021associating} model to track multiple equipments (objects) over time in each camera independently. AOT is a deep learning framework that leverages transformer architectures for accurately associating and tracking objects across video frames.
Given the segmentation mask $\mathbf{M}^{(c)}_{i,t-1}$ from the previous frame and the current frame $\mathbf{F}_{i,t}$, the AOT model $\mathcal{T}$ outputs the tracked mask result $\mathbf{R}^{(c)}_{i,t}$ for object class $c$ at time $t$:
\begin{equation}
\mathbf{R}^{(c)}_{i,t} = \mathcal{T}(\mathbf{M}^{(c)}_{i,t-1}, \mathbf{F}_{i,t}).  
\end{equation}

This process ensures the consistent motion trajectory and accurate mask of each equipment index over time for subsequent collision detection.

\subsection*{2.2.3 Geometric Analysis for Realtime Collision Monitoring} 

Collision monitoring aims to detect the minimum distance of each equipment and warn when it is less than a threshold. In this way, our system will provide autonomous protection of collision. To detect potential collisions between equipment/objects, the system computes the minimum geometric distance between the contours of different equipment. Given the images in the $i$-th camera at the time $t$, let $\mathbf{C}_1$ and $\mathbf{C}_2$ represent the contours of two equipment in the $i$-th camera, and let $\mathbf{p} \in \mathbf{C}_1$ and $\mathbf{q} \in \mathbf{C}_2$ be the points on these contours. The minimum distance between these contours is defined as the minimum Euclidean distance between any pair of points from the two contours: 
\begin{equation}
d_{i,t}(\mathbf{C}_1, \mathbf{C}_2) = \min_{\mathbf{p} \in \mathbf{C}_1, \mathbf{q} \in \mathbf{C}_2} \| \mathbf{p} - \mathbf{q} \|_2.     
\end{equation}

However, computing all points incurs a high computational cost, which affects the real-time performance of the system. To reduce the computational load for real-time processing, we only sample the sparse points $\mathbf{p}, \mathbf{q}$ in the contours $\mathbf{C}_1$ and $\mathbf{C}_2$. By calculating the minimum distance between the contours in this way, the system can determine if two objects are close enough to trigger a potential collision monitoring. Because of the self-occlusions of multiple equipment in different view angles, a collision warning will be triggered if every view finds a collision. Specifically, if the minimum distance between any equipment classes $c_1$ and $c_2$ is below a threshold $\delta$ across all $N$ camera views, a collision is imminent at the time $t$, which is defined as 

\begin{equation}    
\mathcal{W}(c_1, c_2, t) = 
\begin{cases} 
1, & \text{if } d_{i,t}(c_1, c_2) \leq \delta, \forall i \in \{1, 2, \ldots, N\} \\
0, & \text{otherwise}.
\end{cases}
\end{equation}

\subsection*{2.2.4 Transfer Learning}

Because the high-precision equipment in the real beamline environment is very expensive, it is not appropriate to directly develop and test the proposed computer vision method in the real beamline environment.   

\begin{figure}[!t]
\centering
\includegraphics[width=0.5\columnwidth]{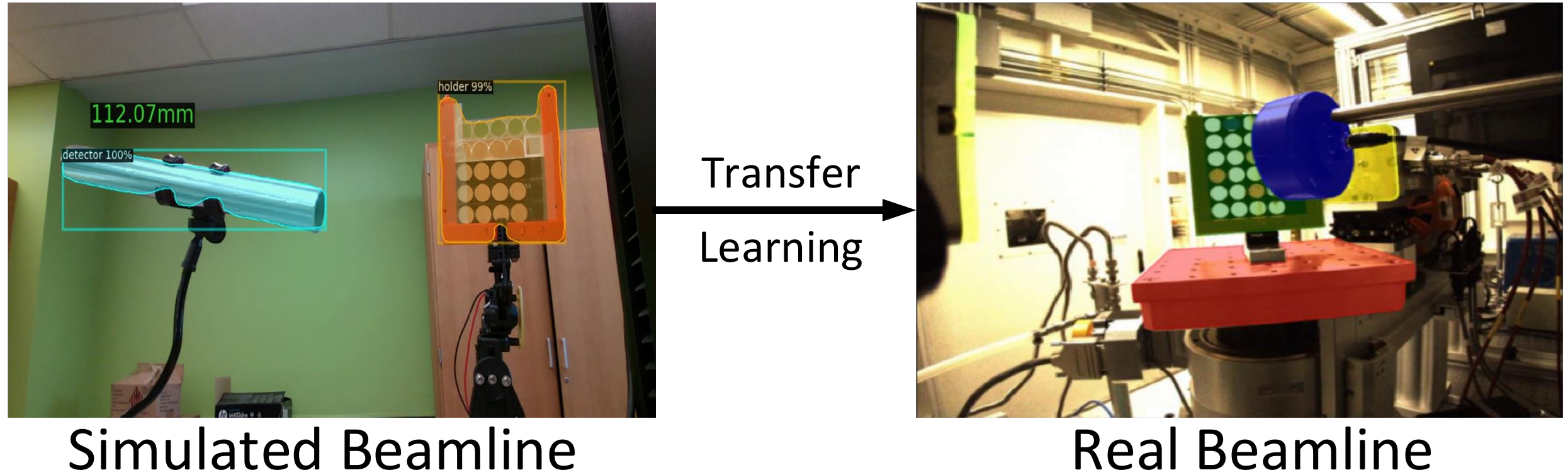}%
\caption{\textbf{Transfer Learning from Simulated Beamline to Real Beamline.} Left: simulated beamline platform at Cleveland State University, Right: real beamline environment at BNL.}
\label{fig:transfer learning}
\end{figure}

To solve this concern, we set up a simulation platform which includes some moving objects similar to the real beamline environment with a simple robot arm, as displayed in Fig.~\ref{fig:transfer learning}. We first train the DNN based computer vision models using the simulated beamline data and then fine-tune the trained DNN models using the real beamline data. Transfer learning is a technique where a model pre-trained on one domain (e.g., simulated data) is fine-tuned on another domain (e.g., real data). This approach improves robustness and reduces data requirements for the real environment.
Our transfer learning from the simulated beamline to real beamline has two significant advantages: 1) We can reduce the experiment risks by testing the computer vision based system on the simulated beamline platform first, 2) The simulated beamline platform could generate diverse challenging scenarios to enrich the pre-training data, which enhances the robustness of the DNN based computer vision models by learning from simulated data to real data. 

\subsection*{2.2.5 Interactive Annotation for New Equipment Classes}
The users might bring their own new equipment to conduct  experiments in the real beamline environment at Brookhaven National Laboratory, which cannot be recognized by the already deployed computer vision models to recognize the commonly used equipment classes. To improve the model's generalization to new equipment classes, we develop an interactive function for users to annotate new equipment classes. The user just needs to click the new equipment classes with the mouse several times in a small number of image frames, then our system will automatically extract the new equipment mask and boundary with the Segment Anything Model (SAM)~\cite{kirillov2023segment}. SAM is a foundational computer vision model designed to segment any object within an image, requiring minimal manual input.
Then, the AOT tracking model~\cite{yang2021associating} is used to track the interested equipment to automatically generate its mask and boundary. The software interface as shown in Fig.~\ref{fig:new_class_software} is user-friendly to refine the tracked masks of the new equipment classes by simple mouse clicks. Finally, the DNN based computer vision models can be fine-tuned so as to recognize the previous equipment classes and the new equipment classes defined by the users simultaneously. 

\section*{III. Experiment}

\subsection*{3.1 High-Precision Beamline Configuration} 
The High-Precision beamline equipment serves as the cornerstone of synchrotron radiation facilities, enabling precise control and manipulation of X-ray beams for a variety of scientific applications. The setup typically encompasses an array of optical components, such as monochromators, mirrors, and slits, as well as detectors and sample stages. Each component plays a critical role in steering, focusing, and tunning the X-ray beams for specific experimental needs, laying the foundation for high-accuracy beamline experiments.

For the module of Multi-Equipment Segmentation and Tracking, a deep learning method is designed to recognize and track the multiple beamline equipments in real time, where the geometry prior is used as regularization. Deep learning is a subset of machine learning that utilizes multi-layered neural networks to automatically extract hierarchical features. Here, it enables the recognition and tracking of multiple beamline equipment. For the module of Geometric Analysis for Realtime Collision Monitoring, we will calculate the paired equipment distance based on the segmentation and tracking results. For the module of Feedback, once the distance between any two equipments is smaller than a predefined threshold, the system will give the collision monitoring and stop the hardware movement in the beamline operation. 

\subsection*{3.2 Hardware Devices}

\textbf{Multi-View Cameras}: Multiple cameras are installed to monitor the equipment operation in the beamline environment. Cameras are positioned at optimal vantage points to capture overlapping fields of view, thereby minimizing blind spots and enhancing spatial awareness. This multi-view configuration ensures robust monitoring of complex scenes, particularly in scenarios involving rapid object motion or occlusions. There are three high-resolution cameras installed, and their spatial positions are uniformly distributed in the 3D space (120 degrees for each pair). Three Allied Vision Mako G-319 high-resolution cameras are installed, each equipped with a Sony IMX265 CMOS sensor. These cameras provide an image resolution of 2064 × 1544 pixels and operate at a frame rate of 37.5 fps.

\textbf{Computation Workstation}: In order to achieve the realtime computation for collision monitoring, a workstation of high-performance deep neural network computation is used. The workstation has two NVIDIA 3090 Graphics Processing Units (GPU) to accelerate the deep neural network computation for parallel computing, which makes the realtime collision warning possible.     

\subsection*{3.3 Software}
We develop a software to integrate all the functional modules of the proposed system: Data Collection with Multi-view Cameras, Multi-Equipment Segmentation and Tracking, Geometric Analysis for Realtime Collision Monitoring, and Feedback.

\begin{figure}[!t]
\centering
\includegraphics[width=0.5\columnwidth]{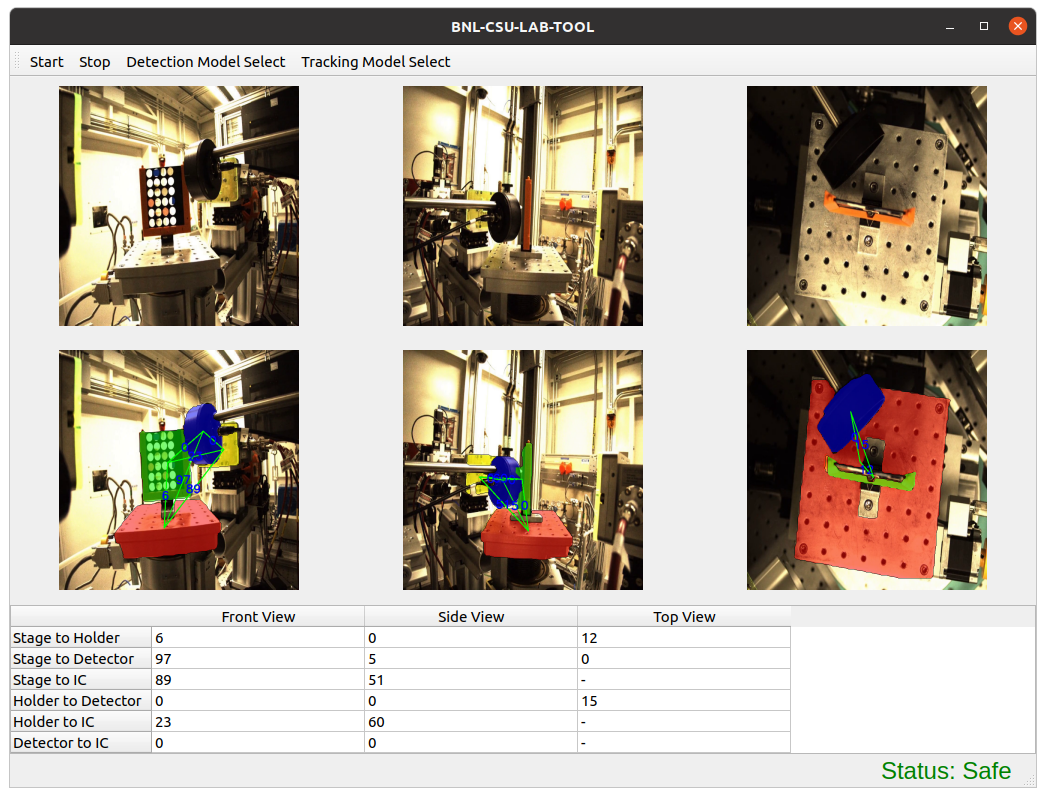}%
\caption{\textbf{Software interface of the proposed system.} From top to bottom: original camera images, segmentation and tracking results, computed equipment distance, system status.}
\label{fig:main_software}
\end{figure}

\textbf{Software Interface}: As shown in Fig.~\ref{fig:main_software}, the developed software interface is clear for the realtime safety monitor. Once initialized, the software interface clearly displays the original camera images, segmentation and tracking results, the computed equipment distance, and the status of the system. The software was developed using Python and PyTorch for the computer vision and deep learning algorithmic components, while the front-end graphical user interface (GUI) was implemented using PyQt. The application is designed to be cross-platform, making it compatible with multiple operating systems,  such as Linux, Windows.

\begin{figure}[!t]
\centering
\includegraphics[width=0.5\columnwidth]{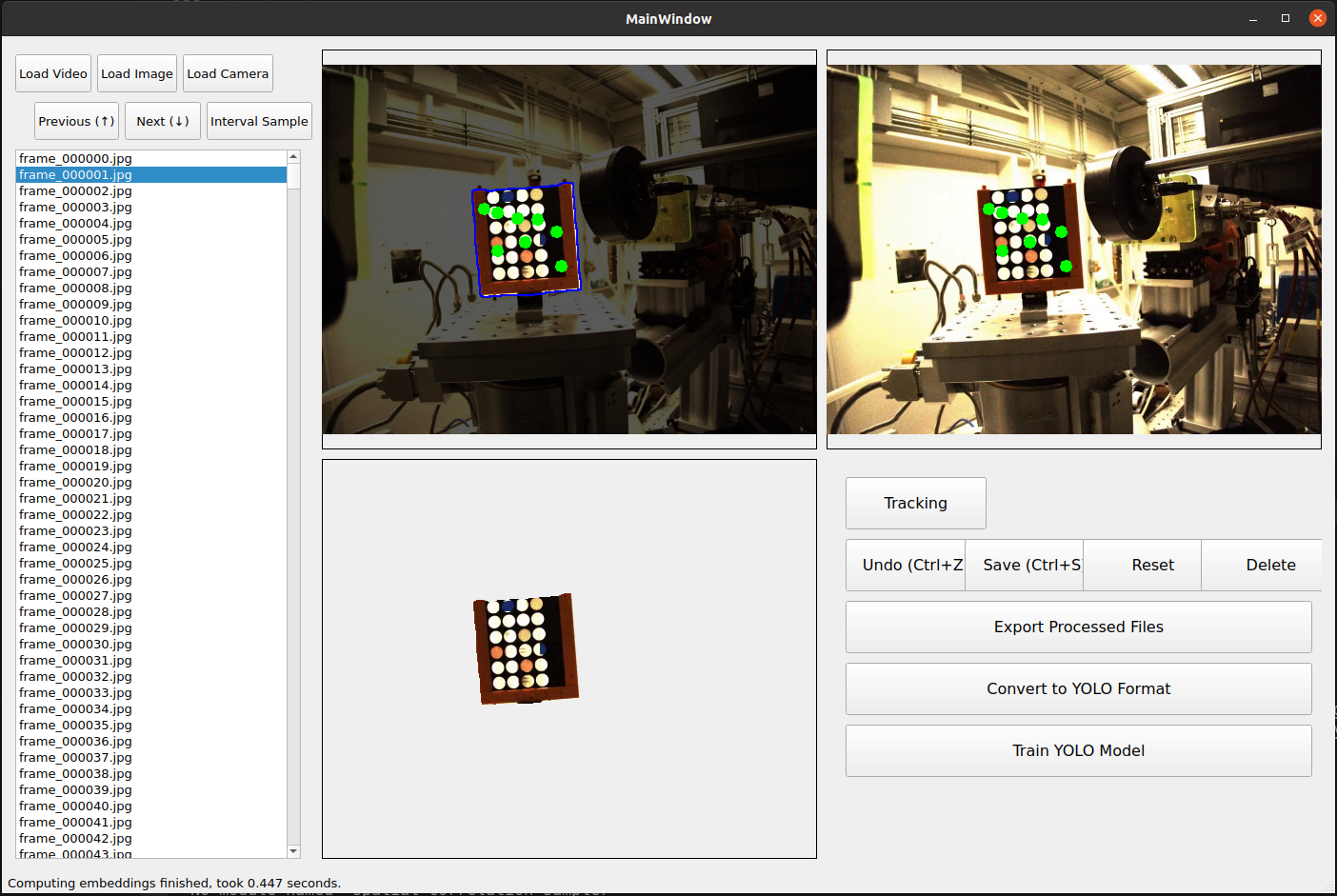}%
\caption{\textbf{Software Interface for Interactive Annotation to New Equipment Classes.} To recognize the new equipment classes not in the previous training set, users could add new equipment classes through simple user mouse clicks.}
\label{fig:new_class_software}
\end{figure}

\textbf{Interactive Annotation for New Equipment Classes}: 
Diverse experimental operations might bring new equipment classes in the beamline environment which are not in the previous training set, so it is necessary to make the system have the capability to recognize the new equipment classes. To achieve this goal, we provide a function of interactive user annotation for new equipment classes as shown in Fig.~\ref{fig:new_class_software}. The user interaction is just as simple as some mouse clicks. After obtaining the annotated new equipment classes, we could fine-tune the existing pre-trained deep learning models to recognize the previous and new equipment classes.

\subsection*{3.4 Dataset}
\textbf{Simulated Beamline Dataset:} We simulated the beamline platform with a fixed simulated Detector and a moving real Holder on a flexible robot arm in a simple controlled environment. We collected a video of 870 frames from the front view, where the two-class object masks are manually annotated by computer vision experts. The DNN based computer vision model will be first trained on the Simulated Beamline Dataset and fine-tuned on the Real BNL Beamline Dataset. This transfer learning from simulated to real data will improve the robustness of our developed computer vision model, as shown in Fig.~\ref{fig:transfer learning}. 

\begin{figure}[!t]
\centering

\includegraphics[width=0.8\columnwidth]{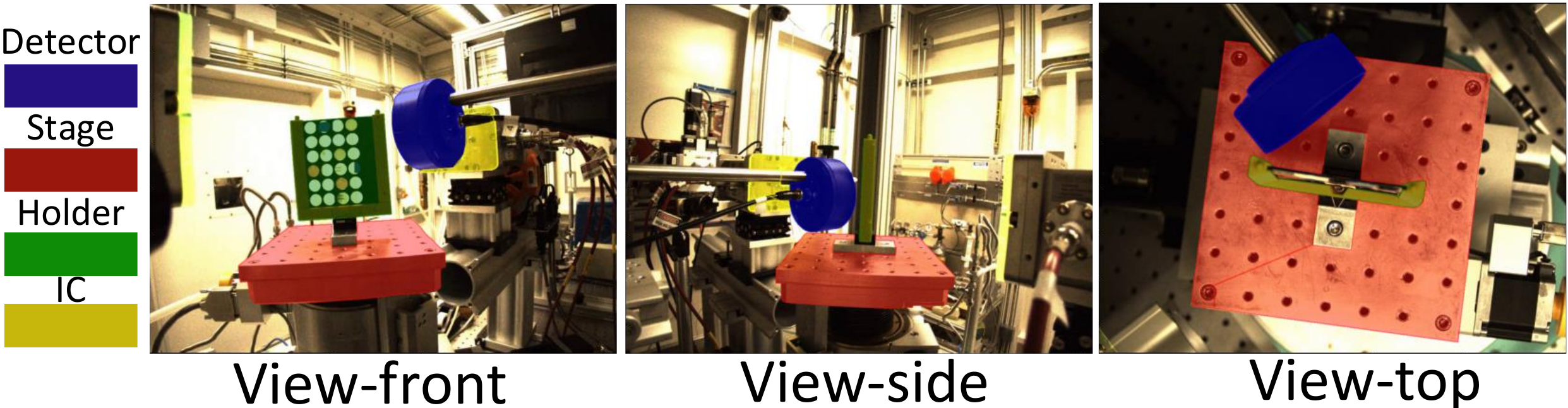}%

\caption{\textbf{Visualization of the Real BNL Beamline Dataset.} Each object class of every frame is labeled with its instance segmentation mask, and the collision monitoring status is also annotated for each frame.}
\vspace{-1em}
\label{fig:dataset}
\end{figure}

\textbf{Real QAS BNL Beamline Dataset:} We collected and annotated three videos (1,100 frames in total) from front, side, and top views in the real beamline environment at the Brookhaven National Laboratory (BNL). The real beamline dataset captures 4 object classes — stage, holder, detector, and IC (ion chamber) —and includes diverse collision and normal scenarios in the real beamline environment. The data was recorded from three different camera views, with externally mounted cameras positioned around the real beamline experimental setup. Each frame is accompanied by detailed object masks manually annotated by computer vision experts to facilitate precise analysis. We further divided dataset into training set and testing set, where training set includes 800 frames (front: 400, side: 200, top: 200) and testing set includes 300 frames (front: 100, side: 100, top: 100). See the Fig.~\ref{fig:dataset} for the visualization examples of the Real BNL Beamline Dataset. 

\subsection*{3.5 Experimental Results} 
In this section, we present the intermediate results of each step in our system, systematically demonstrating the evaluation results of our proposed method. Here, we divide the whole experimental results into four parts: 1) segmentation result, 2) tracking result, 3) distance measurement result, and 4) collision monitoring result. The experiment setting and evaluation metrics are explained in every part separately.

\textbf{Equipment Recognition Results}: The object/equipment instance segmentation aims to draw the mask of multi-class objects within each single frame, and the tracking will smooth the object boundary. To evaluate the equipment recognition performance, we use Intersection over Union (IoU)~\cite{luo2020adversarial} as the primary metric, which measures the overlap between predicted and ground-truth regions for each class. The mean Intersection over Union (mIoU)~\cite{luo2020adversarial} is further calculated by averaging the mask IoU across all classes, providing a comprehensive assessment of segmentation accuracy.

Table.~\ref{tab:seg_result} shows the evaluation result of equipment recognition in different views, which includes four object classes: Holder, Detector, Stage, and IC. The equipment recognition performance is highest in View-top (90.9\% mIoU) and lowest in View-side (88.7\%), which can be attributed to the greater object distance and smaller apparent object sizes in side views, making accurate segmentation more challenging. The Stage consistently achieves the highest IoU over other classes, while the IC class has the lowest, suggesting challenges in distinguishing smaller objects.

\begin{table}[htbp]
\centering
\caption{Equipment recognition results (Mask IoU) across multiple views.}
\label{tab:seg_result}
\begin{tabular}{c|c|c|c|c|c}
\hline
 & \textbf{Holder} &\textbf{Detector} & \textbf{Stage} & \textbf{IC} & \textbf{mIoU} \\ \hline
View-front    &  93.2    &   93.6   &   92.4   &   81.5    &        90.2   \\ \hline
View-side      &    81.9 &    93.5     &           94.6     &        84.8      &  88.7             \\ \hline
View-top     &    85.0         &   92.4                &     95.3           &     -      &     90.9          \\ \hline
\end{tabular}
\end{table}

\textbf{Distance Measurement Result}:
The distance measurement aims to get the pixel-wise distance between any two objects, which provides quantitative results supporting collision monitoring. Here, we use Mean Absolute Error (MAE) and its  
Standard Deviation (STD) as metrics to evaluate the error of the distance calculation. These metrics ensure a robust assessment of the system's capability to estimate distances accurately and reliably.

Table.~\ref{tab:distance_result} reports the pixel-wise distance calculation results for different object pairs across multiple views. H2D is Holder to Detector, H2S is Holder to Stage, H2I is Holder to IC, D2S is Detector to Stage, D2I is Detector to IC, S2I is Stage to IC. The results show that all the views have relatively small distance measurement errors, for example, all the MAE errors are less than 3.9 pixels. The MAE values indicate that the system maintains a high level of accuracy in distance measurement, which is crucial for reliable collision monitoring. 

\begin{table}[htb]
\centering
\caption{Distance Measurement Errors (MAE / STD) between any two objects. H2D: Holder to Detector, H2S: Holder to Stage, H2I:  Holder to IC, D2S: Detector to Stage, D2I: Detector to IC, S2I: Stage to IC. 0/0 indicates that the two objects are always in contact in that view, resulting in both predicted and ground-truth distances being zero, and thus the MAE and STD are also zero. The $–$ symbol denotes cases where the IC is not visible in the top-view, making the corresponding measurement undefined.}
\label{tab:distance_result}

\begin{tabular}{c|c|c|c|c|c|c}
\hline
 & \textbf{H2D} & \textbf{H2S} & \textbf{H2I} & \textbf{D2S} & \textbf{D2I} & \textbf{S2I} \\ \hline
View-front   & 0.2 / 0.4 &    2.6 / 2.4    & 2.7 / 2.6 &       3.9 / 2.8      &  0.0 / 0.0 & 1.7 / 1.5   \\ \hline
View-side    &    2.1 / 1.6    &   3.4 / 1.7     &    1.5 / 2.7         &      1.2 / 1.4     &  0.0 / 0.0  & 2.0 / 1.6    \\ \hline
View-top    &     1.2 / 0.9     &     1.4 / 1.1     &     -         &    0.0 / 0.0      & -   & -  \\ \hline
\end{tabular}
\end{table}

\begin{table}[htbp]
\centering
\caption{Effect of Multi-View Integration on Collision Detection. f: front view, s: side view, t: top view.}
\label{tab:ablation_multiview}
\begin{tabular}{l|c|c|c|c|c|c|c}
\hline
\textbf{View Combination} 
& f 
& s 
& t 
& \makecell{f+s} 
& \makecell{f+t} 
& \makecell{s+t} 
& \makecell{f+s+t} \\\hline

\textbf{Accuracy (\%)} 
& 81.9 & 84.2 & 70.3 & 89.2 & 79.9 & 87.3 & \textbf{99.8} \\\hline
\end{tabular}
\end{table}

\textbf{Collision Detection Result:} Collision monitoring is designed to determine whether two objects in the scene are at risk of colliding, a critical aspect of ensuring safety in beamline operations. The system calculates the minimum distance between the boundaries of objects and compares it against a predefined safety threshold. 
According to the properties of a Gaussian distribution, the range of $\text{MAE} + 2 \times \text{STD}$ covers approximately 95\% of the data. Based on Table~\ref{tab:distance_result}, the maximum observed MAE is 3.9 and the corresponding standard deviation is 2.8. Therefore, we empirically set the collision threshold to 10 pixels. Specifically, if the distance between any two objects is less than 10 pixels, a collision warning is triggered; otherwise, the system is safe. To evaluate the performance of the collision detection module, we use Accuracy as a metric. The system achieves a high accuracy of 99.8\%. 

Our collision detection accuracy 99.8\% means that two frames might have detection errors every 1,000 frames, which cannot satisfy the reliable long-term collision monitoring. To solve this problem, we can set up additional three cameras to build a parallel system, where the original and parallel systems are two identical but independent systems. By fusing the results of two independent systems, the collision detection error (theoretic probability) is reduced to 4 frames every 1M frames. Furthermore, we can issue a true warning if collisions are detected in two consecutive frames. Then, the collision detection error (theoretic probability) is significantly reduced to 16 frames every 1 trillion frames. Theoretically, with a running speed of 10 FPS, only one frame might have detection errors every 198 years for our system. Therefore, our system can perform long-term reliable collision monitoring with simple adjustment.

\subsection*{3.6 System Running Time}
We evaluated the running time of our segmentation and tracking system using a single NVIDIA RTX 3090 GPU. The system processes data from three camera streams simultaneously, performing segmentation, tracking, distance computation, and collision detection, and GUI visualization. 

To enhance real-time multi-view processing, we employ a multi-threaded pipeline where each camera stream is processed in parallel on a single NVIDIA RTX 3090 GPU card, optimizing GPU utilization and reducing latency. The results are then aggregated across views for efficient collision assessment. Under this configuration, the overall system, including computation and visualization, achieves an average running speed of 6 Frames Per Second (FPS). 6 FPS is achieved by using a single NVIDIA RTX 3090 GPU card to process three cameras by our system. If we use a more powerful GPU card or multiple GPU cards, the system's running speed can be improved to 10 FPS. Because the equipments move slowly in the real beamline environment at BNL, the system running speed of 10 FPS is able to satisfy the requirement of realtime collision monitoring.

\subsection*{3.7 Ablation Studies}
To further evaluate the impact of different components in our system, we conduct ablation studies by systematically removing key components, including tracking and multi-view integration, to observe their effect on the overall collision detection performance.

\textbf{Impact of Tracking:} 
Without tracking, object identities may become inconsistent across frames, resulting in missed detections or false positives. We find that the object mask boundary is smoother and more accurate by integrating tracking, and the overall accuracy is improved from 97.1\% to 99.8\%, ensuring that collision monitoring is reliable for the real-world beamline safety.

\textbf{Impact of Multi-View Integration:}  Table~\ref{tab:ablation_multiview} highlights the effectiveness of multi-view integration. Using only a single camera significantly limits the collision monitoring performance due to occlusions and a limited field of view. Combining all three views achieves the highest accuracy (99.8\%), demonstrating that multi-view fusion substantially enhances the robustness of collision detection.

These ablation studies validate that both tracking and multi-view integration play critical roles in ensuring the accuracy and reliability of our proposed autonomous beamline safety system.

\subsection*{3.8 Visualized Results}
\begin{figure*}[!t]
\centering

\includegraphics[width=1\textwidth]{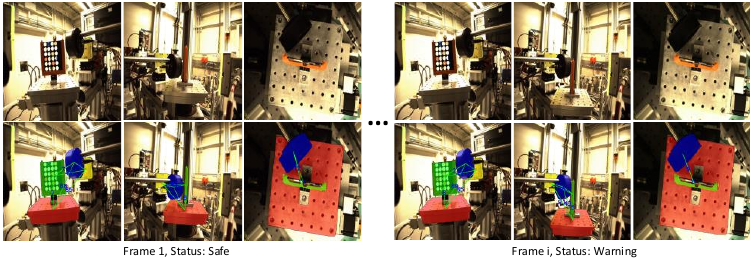}%

\caption{\textbf{Visualization of our system running.} Outcomes of multi-object segmentation, tracking, and collision detection across multiple camera views.}
\vspace{-1em}
\label{fig:vis}
\end{figure*}

Our developed system also provides visualized results to demonstrate the accurate multi-object segmentation, tracking, and collision monitoring in the real beamline environment. Fig.~\ref{fig:vis} illustrates the segmentation results from multiple camera views, showing the clear boundaries of key objects, including the Holder, Detector, Stage, and IC. Fig.~\ref{fig:vis} presents the visualized segmentation and tracking results over consecutive frames, highlighting the consistency in object identity maintenance despite object movement and occlusion. 

For the collision monitoring, Fig.~\ref{fig:vis} visualizes the calculated minimum distances between each object pair and shows whether the collision monitoring is triggered or not. These visualizations confirm that our system effectively captures real-time object interactions and ensures the beamline operation safety. The detailed visual analysis provides intuitive validation for the quantitative metrics reported earlier, showcasing the robustness and reliability of the proposed approach in real-world beamline environments.

\section*{V. Conclusion}
In this paper, we proposed a computer vision-based system to transit the automated beamline into a fully autonomous beamline. By integrating multi-view cameras with computer vision algorithms of object instance segmentation and multi-object tracking, our system achieves real-time collision detection with high accuracy, ensuring safety and efficiency in real-world beamline operations. Additionally, the incorporation of a new class annotation module enhances adaptability by enabling the detection of new object classes. Experimental results on a diverse real-world beamline dataset demonstrate the system's robustness and scalability. This work provides a foundation for achieving fully autonomous beamline operations, paving the way for the the next generation of autonomous synchrotron radiation facilities.

\section*{VI. Acknowledgment}
This research used 7-BM and 8-BM of the National Synchrotron Light Source II, a U.S. Department of Energy (DOE) Office of Science User Facility operated for the DOE Office of Science by Brookhaven National Laboratory under Contract No. DE-SC0012704.

\bibliography{NC25}

\end{document}